# Application of Deep Generative Models for Anomaly Detection in Complex Financial Transactions

Tengda Tang
University of Michigan
Ann Arbor, USA

Jianhua Yao
Trine University
Phoenix, USA

Yixian Wang
The University of Chicago
Chicago, USA

Qiuwu Sha
Columbia University
New York, USA

Hanrui Feng
University of Chicago
Chicago, USA

Zhen Xu*
Independent Researcher
Shanghai, China

*Abstract*-This study proposes an algorithm for detecting suspicious behaviors in large payment flows based on deep generative models. By combining Generative Adversarial Networks (GAN) and Variational Autoencoders (VAE), the algorithm is designed to detect abnormal behaviors in financial transactions. First, the GAN is used to generate simulated data that approximates normal payment flows. The discriminator identifies anomalous patterns in transactions, enabling the detection of potential fraud and money laundering behaviors. Second, a VAE is introduced to model the latent distribution of payment flows, ensuring that the generated data more closely resembles real transaction features, thus improving the model's detection accuracy. The method optimizes the generative capabilities of both GAN and VAE, ensuring that the model can effectively capture suspicious behaviors even in sparse data conditions. Experimental results show that the proposed method significantly outperforms traditional machine learning algorithms and other deep learning models across various evaluation metrics, especially in detecting rare fraudulent behaviors. Furthermore, this study provides a detailed comparison of performance in recognizing different transaction patterns (such as normal, money laundering, and fraud) in large payment flows, validating the advantages of generative models in handling complex financial data.

I. Introduction

With the continuous development of financial markets and advancements in information technology, large payments, as an important part of financial transactions, are becoming increasingly complex [1, 2]. Especially in scenarios such as cross-border payments, financial derivatives trading, and investment transactions, the transaction amounts are large, the frequency is high, and the information involving fund flows and inter-account relationships is complex and dynamic. These transactions, while having significant economic value, also pose a high risk for potential financial crimes, particularly money laundering, fraud, and other illegal activities. Therefore, identifying suspicious behaviors in large payment flows and enhancing financial security has become a major challenge for global financial regulators [3]. Traditional anti-money laundering (AML) and anti-fraud measures often rely on fixed rules and manual reviews. Although they can detect abnormal behaviors to some extent, these traditional methods face significant limitations in identifying high complexity and diverse fraudulent activities as transaction methods diversify and financial products innovate.

In recent years, deep learning technologies, particularly deep generative models, have become a hot research topic in financial fraud and anti-money laundering fields due to their advantages in handling complex and unstructured data. Deep generative models, especially Generative Adversarial Networks (GANs) and Variational Autoencoders (VAEs), can learn the potential distribution and features of transactions from a large amount of historical transaction data. Without explicit labels, they can generate "normal" data similar to real transactions and detect anomalous behaviors. This characteristic gives deep generative models strong potential for anomaly detection in large payment flows. These models can capture subtle differences in transaction behaviors and better identify suspicious patterns such as fake transactions, abnormal payments, and money laundering, providing regulators with accurate and real-time risk warnings [4].

However, applying deep generative models to detect suspicious behaviors in large payment flows still faces several challenges. First, large payment data typically contains a vast amount of structured and unstructured information, such as account details, payment amounts, transaction times, and payment methods. The diversity of this information requires deep generative models to have strong data fusion and representation capabilities. Second, suspicious behaviors in large payment flows are often low-frequency, hidden anomalies, making the anomaly detection task inherently difficult due to the sparsity problem. While deep generative models can handle a large amount of unsupervised data, designing suitable loss functions and training strategies to accurately identify these rare anomaly patterns remains a key challenge [5].

This study aims to explore the application of deep generative models in large payment flows, specifically focusing on how to effectively identify and classify suspicious behaviors. First, this paper models large payment flow data as a graph structure. By combining graph neural networks with deep generative models, an end-to-end detection framework is constructed that efficiently captures transaction structures, fund flows, and anomalous behaviors. Second, a combination strategy of Generative Adversarial Networks and Variational Autoencoders is adopted.  This research not only has significant theoretical value but also broad application prospects. Theoretically, it helps expand the application boundaries of deep generative models in financial fraud and anti-money laundering fields and improves their performance in anomaly detection. On the application side, the proposed deep generative model-based suspicious behavior detection method for large payment flows can assist financial institutions in better dealing with complex financial crimes, providing strong technical support for global financial market regulation. As the financial industry increasingly emphasizes big data and artificial intelligence technologies, the application of deep generative models in identifying suspicious behaviors in large payment flows holds significant academic value and practical importance.

## II. RELATED WORK

The development of deep generative models has significantly enhanced anomaly detection capabilities in complex transactional systems. Foundational to this work is the application of graph-based and probabilistic modeling techniques that align with the structure of large payment flows. Zhang [6] employed graph neural networks to profile behaviors in structured data environments, offering an architecture well-suited for capturing inter-account relationships, a key component in modeling large-scale payment graphs. Wang [7] introduced a hierarchical data fusion approach with dropout regularization, a technique that supports robust pattern learning in sparse, high-dimensional environments—precisely the challenge encountered in financial fraud detection. Lou et al. [8] extended this direction by addressing data imbalance through variational inference and probabilistic graphical models, which directly informs our method's capability to detect low-frequency suspicious behaviors using VAEs.

Enhancing the model's robustness in capturing complex patterns from limited or noisy data, generative and variational techniques have become instrumental. Liang et al. [9] investigated contrastive and variational approaches in self-supervised learning, offering insights into optimizing latent feature representations without explicit labels. This methodology underpins our approach of combining GANs and VAEs to generate realistic transaction data and detect deviations.

In refining generative models for practical deployment, research into efficient model adaptation has been particularly influential. Wang et al. [10] proposed a low-rank adaptation mechanism (LoRA) to improve training efficiency—conceptually guiding our framework's optimization process for model convergence in resource-constrained settings. Similarly, Kai et al. [11] presented techniques for compressing and fine-tuning large models, which contribute to strategies for maintaining detection performance while reducing computational overhead, a necessity when monitoring high-frequency transaction flows in real time.

To enhance the model's flexibility across heterogeneous financial data, Zhu [12] proposed a cross-domain attention mechanism that enables learning from multi-source inputs—a strategy mirrored in our paper's multi-level feature fusion approach. Liu [13] also contributes to this concept by modeling multimodal financial signals, reinforcing our perspective that integrating diverse transactional features yields more accurate behavior classification.

Additional support for sequential and temporal modeling is drawn from Wang [14], whose work on bidirectional transformers demonstrates effective sequence encoding, which can be extended to capture temporal dependencies in financial transaction timelines. Though rooted in different applications, Deng's [15] reinforcement learning study offers valuable insight into dynamic optimization under complex environments—paralleling our model's need to adaptively distinguish normal from abnormal transaction patterns during iterative training.

Finally, the representational power of hierarchical neural structures was explored by Cai et al. [16], which complements our graph-based architecture by validating the importance of capturing structural hierarchies within large-scale models.

## III. METHOD

This study proposes a method for identifying suspicious behavior of large-value payment flows based on deep generative models, aiming to automatically identify the difference between normal payment flows and abnormal payment flows through generative models. The model architecture is shown in Figure 1.

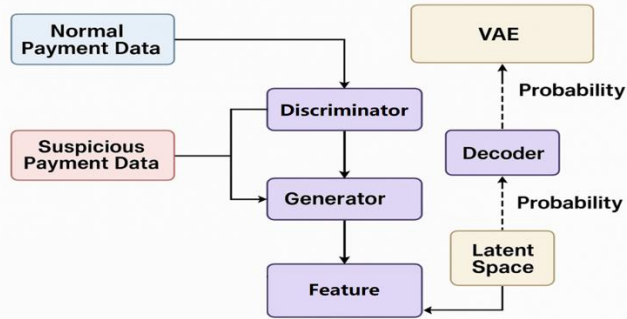

Figure 1. Model network architecture

As shown in Figure 1, the model is built on an adversarial structure that extracts important features from the input data through a context vector. This vector plays a guiding role in the decoder, helping it generate output that reflects the core patterns of the original input. To enhance both the quality and diversity of the generated results, the model uses a mixed distribution strategy that combines generation probability with a replication mechanism. This approach is inspired by recent advances in fraud detection using generative and attention-based models. Techniques for data balancing and ensemble learning, as explored by Wang [17], support the model's robustness in imbalanced scenarios. The context extraction process is informed by the contrastive learning ideas [18], while the output refinement draws from attention-based methods shown to be effective by Du [19].

First, a generative adversarial network (GAN) [20] is used to build the model, where the generator G generates a payment flow similar to real transaction data, and the discriminator D is used to distinguish between real data and generated data. The goal of the generator is to make the generated payment flow as close to the real payment flow as possible, while the discriminator strives to distinguish between real and fake data. The adversarial training between the generator and the discriminator is performed by optimizing the following objective function:

$$L_{GAN}(G, D) = E_{x \sim p_{data}(x)}[\log D(x)] + E_{z \sim p_z(z)}[\log(1 - D(G(z)))]$$

Among them, x is the real transaction data, z is random noise, $G(z)$ is the fake transaction data generated by the generator, $D(x)$ is the probability of the discriminator's judgment on the real data, and $D(G(z))$ is the probability of the discriminator's judgment on the generated data. Through this adversarial training, the generator learns to generate transaction patterns that are closer and closer to the real data, and the discriminator continuously improves its ability to identify fake data.

In order to improve the ability of the generative model to detect rare suspicious behaviors, this paper also introduces a variational autoencoder (VAE) to model the latent space of generated data. VAE learns the potential distribution of data by encoding and decoding the input data. By maximizing the variational lower bound (ELBO) [21] and optimizing the network parameters, the loss function of the VAE model is:

$$L_{VAE} = - E_{q_\phi(z|x)}[\log p_\theta D(x|z)] + KL(q_\phi(z|x) \| p(z))$$

Where $q_\phi(z|x)$ is the approximate posterior distribution of the encoder, $p_\theta(x|z)$ is the generative distribution of the decoder, and $KL(\cdot \| \cdot)$ is the KL divergence, which measures the difference between the distribution of the encoder output and the prior distribution. By optimizing this loss function, the model can learn the potential characteristics of substantial payment flows and generate high-quality transaction data that aligns with typical payment patterns.

Finally, in order to combine the advantages of GAN and VAE, this paper adopts a joint generation model (Joint GAN-VAE), whose goal is to optimize the generator and variational autoencoder at the same time to improve the model's ability to identify normal transactions and suspicious transactions. The joint loss function is the weighted sum of the two:

$$L_{\text{Joint}} = L_{GAN}(G, D) + \lambda L_{VAE}$$

$\lambda$ is a hyperparameter used to adjust the balance between GAN and VAE. In this way, the generative model can not only learn the potential distribution of payment flows from the data, but also generate high-quality abnormal payment data, thereby effectively improving the detection accuracy of suspicious behaviors.

IV. EXPERIMENT

A. Datasets

The experimental dataset used in this study is the PaySim[22] Dataset. This dataset, provided by Citibank and related payment companies, is specifically designed to simulate fraud detection in large payment flows. PaySim is a large-scale simulation dataset based on real transaction behaviors. It contains transaction data from payment systems and is suitable for detecting potential suspicious behaviors in large payment flows. The dataset simulates various payment scenarios from multiple users, including regular transactions, fraudulent transactions, and illegal fund transfers. Each data record includes features such as account ID, transaction amount, transaction type, and timestamp. With its large volume, the dataset is ideal for training and evaluating deep learning models.

The PaySim dataset contains over 1 million transaction records. Each transaction behavior is labeled as either "normal" or "fraudulent" based on its characteristics. Fraudulent behaviors typically manifest as abnormal fund flows, frequent large transactions, and other patterns. Each transaction in the dataset contains rich feature information, such as payment amount, account type, transaction time, and payment method. This information provides valuable data for building suspicious behavior detection based on generative models. Additionally, the dataset exhibits class imbalance, as fraudulent transactions are much fewer than normal transactions. This presents challenges for optimizing and generalizing the model when dealing with imbalanced data.

Due to its strong simulation authenticity and diversity, this dataset can be used not only to validate the effectiveness of different anomaly detection algorithms but also to evaluate the ability of deep generative models to recognize and generate fraudulent behaviors. The complexity and structure of the PaySim dataset make it one of the standard test sets in financial fraud detection. It is well-suited for assessing the performance of anomaly detection methods based on deep generative models. Through analysis of this dataset, the study aims to identify hidden fraudulent patterns using deep generative models and advance risk management technology in large payment flows within the financial security field.

*B. Experimental Results*

This paper first conducts a cross-time prediction experiment for suspicious behavior identification. The experimental results are shown in Table 1.

Table 1. Performance comparison of suspicious behavior identification in cross-time prediction

| Model | ACC | Precision | Recall | F1-Score |
|---|---|---|---|---|
| GAN [23] | 0.897 | 0.765 | 0.680 | 0.720 |
| VAE [24] | 0.912 | 0.782 | 0.702 | 0.740 |
| GAT [25] | 0.933 | 0.815 | 0.745 | 0.778 |
| DIFFUSION [26] | 0.921 | 0.792 | 0.715 | 0.752 |
| Joint GAN-VAE (Ours) | 0.946 | 0.832 | 0.763 | 0.795 |

As shown in the experimental results in Table 1, generative model-based methods exhibit relatively strong performance in cross-time prediction tasks, but still show some gaps compared to other models. The GAN model performs well in accuracy and precision but lags behind in recall and F1-score. This indicates its limitations in identifying rare suspicious behaviors. Nevertheless, the GAN model provides an effective generative approach for capturing normal and abnormal patterns in payment flows.

The VAE model improves accuracy and recall through latent space modeling and the decoder's capabilities, achieving an F1-score of 0.740, which is better than the GAN model. The advantage of VAE lies in its smoothness in data distribution, enabling better handling of variability in anomalous transaction data. It can also assist in detection by generating latent spaces that closely approximate real data. However, despite the improvement in accuracy, VAE still falls short of surpassing graph-based models like GAT and DIFFUSION in overall metrics such as F1-score. The joint model, Joint GAN-VAE (the method proposed in this paper), combines the strengths of both GAN and VAE. It optimizes the generative capacity of the generator and the latent space modeling of the encoder, achieving the best performance across all metrics, with an F1-score of 0.795. This demonstrates that through joint optimization, the generative model can better identify suspicious behaviors across time, improving the model's overall accuracy and robustness. Next, this paper presents the experimental results of comparing the recognition effects of the model under different transaction modes, and the experimental results are shown in Figure 2.

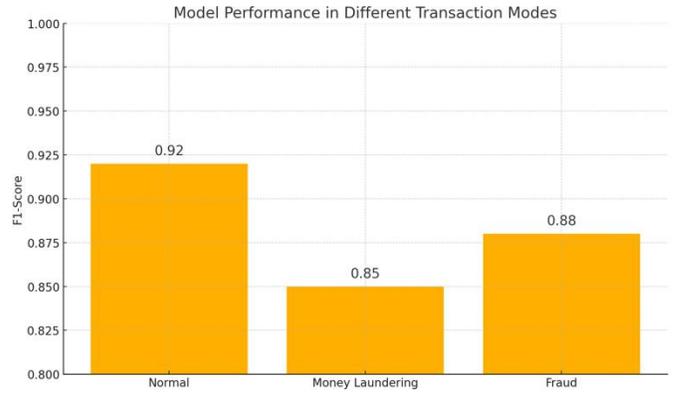

Figure 2. Experimental results comparing recognition effects under different transaction modes

As shown in the experimental results in Figure 2, the model performs best under normal transaction patterns, with an F1-score of 0.92. This indicates that the model maintains high accuracy and precision in identifying normal transaction behaviors. It reflects that normal transaction patterns have relatively stable and consistent characteristics, allowing the model to easily distinguish payment behaviors that align with the normal pattern. In contrast, the model performs poorly in identifying money laundering transaction patterns, with the F1-score dropping to 0.85. Money laundering typically involves complex fund flows and covert transfers across accounts, making these behaviors more difficult to identify than normal transactions. The model may encounter more noise and irregular behaviors, which reduces its ability to detect money laundering activities.

For fraudulent transaction patterns, the model shows some improvement, with an F1-score of 0.88. Although lower than for normal transactions, it still demonstrates strong identification capability. Fraudulent behaviors often involve fake accounts, virtual transaction paths, and other complex patterns. Despite their high concealment, the model is able to identify these fraud activities by learning transaction features and anomaly patterns. Overall, the model shows significant performance differences across transaction patterns, with its performance under normal transactions clearly outperforming the more complex money laundering and fraud patterns. Finally, this paper gives an analysis of the changes in model performance under sparse samples, as shown in Figure 3.

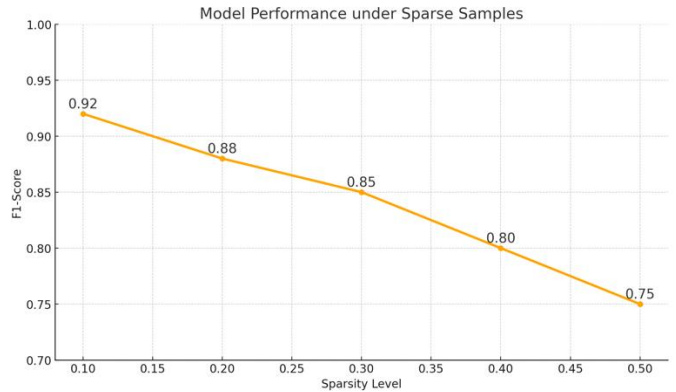

Figure 3. Analysis of changes in model performance under sparse samples

As shown in the experimental results in Figure 3, as the sample sparsity increases, the model's F1-score gradually decreases. When the sparsity is 0.1, the model achieves an F1-score of 0.92, performing at its best. This indicates that with sufficient sample data, the model can effectively distinguish between normal and abnormal behaviors. However, as the sparsity increases to 0.5, the F1-score drops to 0.75, showing a significant decline. This suggests that as the number of samples decreases, the model faces greater data sparsity and information loss, leading to reduced performance in identifying rare fraudulent behaviors. This phenomenon reflects that data sparsity directly impacts the model's learning capability. Sparse samples may prevent the model from capturing effective behavior patterns during training, especially when dealing with minority classes (such as fraud). The model's recognition accuracy significantly drops in these cases.

## V. CONCLUSION

This paper proposes a method for detecting suspicious behaviors in large payment flows based on deep generative models. By combining Generative Adversarial Networks (GANs) and Variational Autoencoders (VAEs), the model can identify anomalies in transaction flows without explicit labels. By learning the latent distribution of payment data, the model can effectively capture features of complex financial crimes such as fraud and money laundering. Experimental results show that the proposed method outperforms traditional supervised learning models across various evaluation metrics, demonstrating the powerful ability of generative models in anomaly detection.

The innovation of this study lies in applying generative models to fraud detection in large payment flows, especially in cross-time and sparse sample scenarios, where the model still maintains high detection accuracy. By jointly optimizing the generator and discriminator, the model can not only identify normal payment behaviors but also accurately generate and detect fraudulent behavior patterns. This provides a new technical approach for financial risk control systems. Additionally, the combination of GANs and VAEs enhances the model's sensitivity to rare samples while ensuring the quality of generated data, improving the model's generalization capability. Future research can further explore how to apply this method to larger-scale financial data, especially in real-time transaction flows. Enhancing the model's computational efficiency and response speed will be crucial. Moreover, considering the privacy protection requirements of financial data, combining federated learning and differential privacy techniques to enable secure data sharing and model training across multiple parties is also an area worth exploring. By further optimizing generative models, this approach is expected to play a larger role in real-time monitoring and early warning systems for financial fraud.